\documentclass{article}

\usepackage[final]{staix_2026}         

\usepackage[utf8]{inputenc}
\usepackage[T1]{fontenc}
\usepackage{hyperref}
\usepackage{url}
\usepackage{booktabs}
\usepackage{amsfonts}
\usepackage{nicefrac}
\usepackage{microtype}
\usepackage{color}
\usepackage{xcolor}
\usepackage{graphicx}
\usepackage{amsmath,amssymb,amsthm}
\usepackage{multirow}
\usepackage{subcaption}  
\usepackage{algorithm}
\usepackage{algpseudocode} 
\usepackage{xcolor}
\usepackage{hyperref}
\usepackage{url}
\hypersetup{hidelinks}

\newcommand{\R}{\mathbb{R}}
\newcommand{\E}{\mathbb{E}}
\newcommand{\Prob}{\mathbb{P}}
\newcommand{\Hist}{\mathcal{H}}
\newcommand{\V}{\mathcal{V}}
\newcommand{\Rel}{\mathcal{R}}
\newcommand{\Edge}{\mathcal{E}}
\newcommand{\Data}{\mathcal{D}}
\newcommand{\Chain}{\mathcal{S}}
\newcommand{\Time}{\mathcal{T}}
\newcommand{\Loss}{\mathcal{L}}
\newcommand\blfootnote[1]{%
  \begingroup
  \renewcommand\thefootnote{}\footnote{#1}%
  \addtocounter{footnote}{-1}%
  \endgroup
}
\newtheorem{definition}{Definition}
\DeclareMathOperator*{\argmax}{arg\,max}
\DeclareMathOperator{\Softmax}{Softmax}
\DeclareMathOperator{\Softplus}{Softplus}
\DeclareMathOperator{\GELU}{GELU}
\DeclareMathOperator{\Concat}{Concat}
\DeclareMathOperator{\MHA}{MHA}
\DeclareMathOperator{\LayerNorm}{LayerNorm}

\title{GAttNHP: Group Attention Neural Hawkes Process for Extrapolation Reasoning in Temporal Knowledge Graphs}

\author{%
  Xiangni Tian\,$^{1}$\\
  \texttt{tianxiangni@itc.ynu.edu.cn}
  \And 
  Kaixian Yu\,$^{2}$\\
  \texttt{kaixian@insilicom.com}
  \And
  Runpeng Dai\,$^{3}$\\
  \texttt{runpeng@email.unc.edu}
  \AND
  Niansheng Tang\,$^{1, *}$\\
  \texttt{nstang@ynu.edu.cn}
  \And
  Hongtu Zhu\,$^{3, *}$\\
  \texttt{htzhu@email.unc.edu}
}

\begin{document}

\maketitle
\blfootnote{\hspace{-6mm}$^{1}$Yunnan Key Laboratory of Statistical Modeling and Data Analysis, Yunnan University, Kunming, China; $^{2}$Insilicom LLC, Peabody, MA, USA; $^{3}$University of North Carolina at Chapel Hill, North Carolina, USA. $^{*}$Co-corresponding authors: Niansheng Tang (nstang@ynu.edu.cn) and Hongtu Zhu (htzhu@email.unc.edu).}

\begin{abstract}
Temporal Knowledge Graphs (TKGs) record how facts evolve over time, but forecasting future events on a TKG remains difficult for three reasons: (i) long-range temporal dependencies are hard to encode; (ii) events on different chains mutually excite or inhibit one another in ways that snapshot-level models cannot express; and (iii) inter-arrival times are heavy-tailed and statistically sparse, so deterministic time predictors are unreliable. We address these three issues with a single framework, the \textbf{Group Attention Neural Hawkes Process (GAttNHP)}, built around three matched components. First, a self-attention encoder casts each subject--relation chain as a continuous-time point process and captures the lingering excitation of distant history. Second, a semantic soft-grouping module turns globally learnable Hawkes priors into an analytical cross-attention mask, so chains share excitation patterns through their latent group memberships rather than through exhaustive pairwise computation. Third, a Non-Crossing Quantile (NCQ) regression head replaces mean-based time prediction, providing calibrated, monotonically ordered quantile estimates that remain stable under heavy-tailed inter-arrival distributions. On six benchmark TKG datasets, GAttNHP improves over state-of-the-art baselines on both entity prediction and time prediction, and ablations confirm that its largest gains arise on the long-tail event chains where existing models fail most severely.
\end{abstract}

\section{Introduction}
\label{sec:intro}

Knowledge Graphs (KGs) encode human knowledge as triplets $(s, r, o)$, representing subject, relation, and object. Because real-world facts evolve over time, \emph{Temporal Knowledge Graphs (TKGs)} augment each triplet with a timestamp, yielding $(s, r, o, t)$. The central challenge in TKG reasoning is \emph{extrapolation}, namely reasoning beyond the observed time horizon. We focus on two extrapolative tasks that jointly characterize a future event: predicting \emph{what} will happen, known as future link prediction $(s, r, ?, t)$, and predicting \emph{when} it will happen, known as occurrence-time estimation $(s, r, o, ?)$. The latter task remains much less explored, despite its importance for downstream systems that must act on forecasts. 
 
{\color{red}\begin{figure}[t]
    \centering
    \begin{subfigure}{0.48\textwidth}
        \centering
        \includegraphics[width=\linewidth]{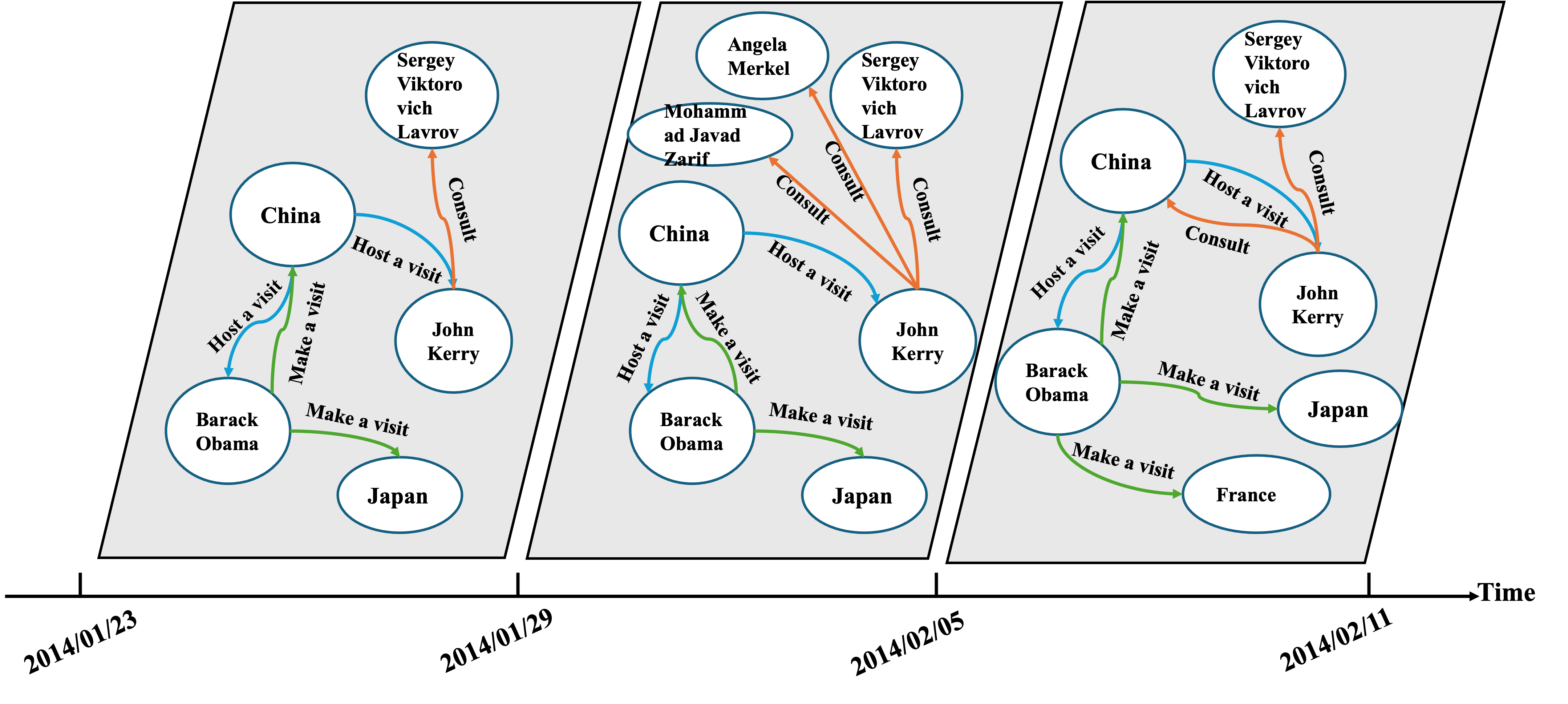}
        \caption{TKG as timestamped relational snapshots.}
    \end{subfigure}\hfill
    \begin{subfigure}{0.48\textwidth}
        \centering
        \includegraphics[width=\linewidth]{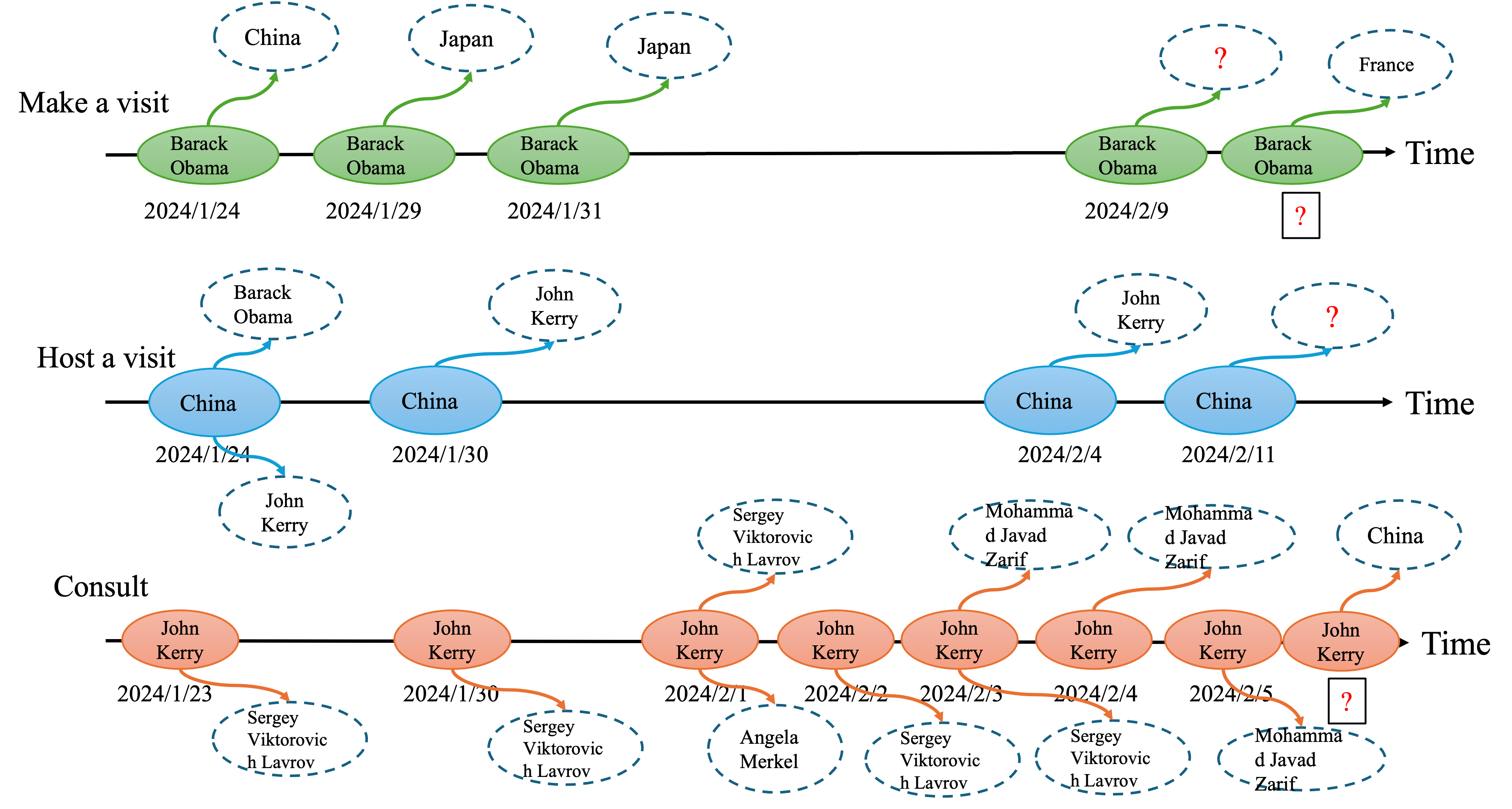}
        \caption{Our view: marked, mutually-exciting event chains.}
    \end{subfigure}
    \caption{From a temporal knowledge graph to a marked point process.
    (a) Conventionally, a TKG is a sequence of timestamped multi-relational
    snapshots; each edge is a fact $(s,r,o)$ such as (Barack Obama, \emph{Make a visit}, China).
    (b) We re-organize the same facts into event chains: each subject--relation pair
    $(s,r)$ forms one continuous-time chain whose timestamped events carry different
    object \emph{marks}. Chains are coupled through shared entities and mutually excite
    one another; ``?'' marks the prediction targets---the next object (future link
    prediction) and its time (occurrence-time prediction).}
    \label{fig:overview}
\end{figure}}

\paragraph{Related work.}
Existing TKG reasoning methods fall into four broad paradigms. \emph{Quad-based} methods such as TTransE~\citep{garcia2018tatranse}, ChronoR~\citep{sadeghian2021chronor}, and ATiSE~\citep{xu2020atise} inject time directly into static entity and relation embeddings. \emph{Graph-based} methods, including RE-NET~\citep{jin2020renet}, 
and CENET~\citep{xu2023cenet}, aggregate structural evolution across temporal snapshots. \emph{Path-based} methods such as TLogic~\citep{liu2022tlogic} and CluSTeR~\citep{li2021cluster} explicitly construct multi-hop reasoning chains. \emph{Transformer-based} architectures, including SToKE~\citep{gao2023stoke} and ECEformer~\citep{fang2024eceformer}, leverage self-attention to capture contextual dependencies. All four paradigms achieve strong performance, yet they share a common limitation: they operate on discrete time steps or static snapshots, oversimplifying the continuous, stochastic nature of real-world events.

Temporal Point Processes (TPPs), and in particular the self-exciting Hawkes process~\citep{hawkes1971spectra,isham1979selfcorrecting,shchur2021neural}, offer a principled probabilistic alternative. Deep extensions of TPPs~\citep{du2016recurrent,mei2017neural,zhang2019selfattentive,zuo2020transformer,yang2022transformer} have shown strong results on generic event streams, and pioneering work, such as Know-Evolve~\citep{trivedi2017knowevolve} GHNN~\citep{han2020GHNN} and the Transformer-integrated GHT~\citep{sun2022ght}, first imported continuous point processes into TKG representation learning. More recent models, such as HTCCN~\citep{chen2024htccn} and THCN~\citep{chen2024thcn}, adapt Hawkes processes to TKG extrapolation through temporal causal convolutions.

Despite this progress, three challenges remain unresolved across both discrete and point-process paradigms:
 \textbf{(i) Insufficient temporal dynamics.} Snapshot-based models ignore the fluid nature of time and struggle to capture the lingering excitation of distant history.
 \textbf{(ii) Neglected mutual excitation.} Most frameworks model entities in isolation and cannot express how the occurrence of one event chain triggers or inhibits another.
 \textbf{(iii) Deterministic, miscalibrated time prediction.} Time estimation has received far less attention than link prediction, and the methods that do exist rely on mean-based regression that is unreliable under heavy-tailed inter-arrival distributions.

\paragraph{Our contribution.}
We propose the \textbf{Group Attention Neural Hawkes Process (GAttNHP)}, a unified framework that addresses these three challenges with three matched components.
 
\noindent (a)  \textbf{Continuous-time self-excitation via attentional point processes.} We cast each subject--relation event chain as a marked TPP, and encode its history with a self-attention mechanism. This bridges discrete snapshots and continuous time, capturing long-range dependencies and dynamically simulating the evolution of the intensity function. \emph{(Addresses challenge (i).)}

\noindent (b) \textbf{Cross-chain mutual excitation via semantic soft grouping.} We dynamically infer a \emph{soft} group membership for each chain, and translate a \emph{globally} learnable, group-level Hawkes prior $(\Phi,\gamma)$ into an analytical additive cross-attention mask. Exponentiated inside the softmax, this mask contributes an exact group-Hawkes excitation--decay factor that \emph{modulates} the semantic query--key similarity, so the model captures mutual excitation among chains and learns shared group-level excitation patterns without exhaustive pairwise computation. \emph{(Addresses challenge (ii).)}

\noindent (c) \textbf{Calibrated time prediction via Non-Crossing Quantile regression.} We replace mean-based regression with an NCQ head that estimates a full set of monotonically ordered quantiles of the inter-arrival distribution. The construction guarantees no quantile crossing and remains stable under heavy tails, yielding both a robust point estimate (the median) and explicit uncertainty quantification. \emph{(Addresses challenge (iii).)}


\noindent 
These components are jointly optimized end-to-end. Section~\ref{sec:background} reviews the Hawkes process. Section~\ref{sec:method} develops GAttNHP. Section~\ref{sec:experiments} presents experiments on TKG benchmarks. Section~\ref{sec:conclusion} concludes.


\section{Background: Hawkes Processes}
\label{sec:background}

A \textbf{Hawkes process}~\citep{hawkes1971spectra} is a doubly stochastic point process whose conditional intensity is
\begin{equation}
\lambda(t) = \mu + \sum_{i:\,t_i < t} g(t - t_i),
\label{eq:hawkes}
\end{equation}
where $\mu \geq 0$ is the base intensity and $g(\cdot)$ is a non-negative, decaying kernel (e.g., exponential or power-law). Each past event lifts the instantaneous likelihood of future events, and that influence decays with time.

A \textbf{multivariate Hawkes process} generalizes Eq.~\eqref{eq:hawkes} to $K$ event types:
\begin{equation}
\lambda_k(t) = \mu_k + \sum_{j=1}^{K} \sum_{t_{j\ell} \in \Hist_{j,t}} g_{kj}(t - t_{j\ell}), \qquad k = 1, \ldots, K,
\label{eq:multivariate-hawkes}
\end{equation}
where $\mu_k$ is the baseline intensity for type $k$, $\Hist_{j,t}$ is the history of type-$j$ events before $t$, and $g_{kj}(\cdot)$ is the excitation kernel describing how past type-$j$ events influence the intensity of type $k$. A common choice is the exponential kernel $g_{kj}(u) = \alpha_{kj} e^{-\beta_{kj} u}  \mathbf{1}\{u > 0\}$, in which $\alpha_{kj} \geq 0$ controls the excitation magnitude and $\beta_{kj} > 0$ governs the decay rate.

The classical Hawkes formulation has two limitations relevant to TKGs. First, the kernel only allows past events to \emph{excite}, never inhibit, future occurrences. Second, parametric kernels are too rigid to capture the rich, context-dependent dynamics of real event streams. \textbf{Neural Hawkes processes} address both limitations by parameterizing the conditional intensities with neural networks. For an event sequence on $(0, T]$, the total intensity is
\begin{equation}
\lambda(t) = \sum_{k=1}^{K} \lambda_k(t) = \sum_{k=1}^{K} f_k\!\left(w_k^\top h(t)\right), \qquad t \in (0, T],
\end{equation}
where $h(t)$ is a continuous-time hidden state summarizing event history, $w_k$ is a learnable weight vector for type $k$, and $f_k(\cdot)$ is a non-negative activation. A common choice is the temperature-scaled softplus $f_k(c) = \tau_k \log(1 + \exp(c/\tau_k))$ with $\tau_k > 0$. We adopt this neural formulation as the foundation of GAttNHP.

\section{Methodology}
\label{sec:method}

This section develops GAttNHP. Section~\ref{subsec:problem} formalizes TKG extrapolation as a marked point process. Section~\ref{subsec:gnhp} then specifies the conditional intensity as the superposition of three components---base, self-excitation, and group excitation---that the rest of the section operationalizes: the self-excitation term is realized in Section~\ref{subsec:self} by an attention-based history encoder, the group excitation term in Section~\ref{subsec:group} by semantic soft grouping with an analytical Hawkes mask, and the full intensity is assembled in Section~\ref{subsec:intensity}. Section~\ref{subsec:ncq} then introduces the Non-Crossing Quantile head for time prediction, and Section~\ref{subsec:training} states the joint training objective.

\subsection{Problem formulation}
\label{subsec:problem}

\begin{definition}[TKG]
A TKG is a sequence of knowledge graphs $\mathcal{G} = \{G_1, G_2, \ldots, G_T\}$, where each snapshot at timestamp $t$ is a directed multi-relational graph $G_t = (\V, \Rel, \Edge_t)$. Here $\V$ is the set of entities, $\Rel$ is the set of relation types, and $\Edge_t$ is the set of facts observed at time $t$. Each fact is a quadruple $(s, r, o, t)$, indicating that subject $s$ is linked to object $o$ by relation $r$ at time $t$.
\end{definition}

\paragraph{TKG reasoning as a marked point process.}
Given an observed dataset $\Data = \{(s, r, o, t)\}$, we associate one event chain $\Chain_u$ with each subject--relation pair $u = (s, r)$. The chain consists of object--time marks $(o_i, t_i)$, and its history up to time $t$ is $\Hist_t^u = \{(o_i, t_i) \in \Chain_u : t_i < t\}$. We model $\Chain_u$ as a marked TPP with conditional intensity
\begin{equation}
\lambda(o, t \mid s, r) = \lim_{\Delta \downarrow 0}\frac{\Prob\!\left(\text{event with mark } o \text{ in } [t, t+\Delta) \mid \Hist_t^u\right)}{\Delta},
\end{equation}
i.e., the instantaneous rate of observing object $o$ at time $t$ for the chain $u$, given the past.

\paragraph{Downstream tasks.}
Once the intensity is learned, we consider two standard TKG reasoning tasks.
\textbf{(i) Entity prediction at time $t$.} Given a query $(s, r, ?, t)$, infer the missing object as
\begin{equation}
\hat{o} = \argmax_{o \in \V} \lambda(o, t \mid s, r, \Hist_t).
\label{eq:entity-prediction}
\end{equation}
\textbf{(ii) Time prediction for a given fact.} Given $(s, r, o, ?)$, forecast the next occurrence as
\begin{equation}
\hat{t} = \E\!\left[T \mid T > t_{\text{now}},\, s, r, o, \Hist_{t_{\text{now}}}\right].
\label{eq:time-prediction}
\end{equation}

\subsection{Group neural Hawkes intensity}
\label{subsec:gnhp}

Following the Group Network Hawkes Process(GNHP)~\citep{fang2024gnhp}, we decompose the conditional intensity for a chain $u = (s, r)$ as
\begin{equation}
\lambda_u(t) = \phi\!\Big(\,\underbrace{\mu_u(t)}_{\text{base}}\;+\;\underbrace{\nu_{\text{self}}(t, \Hist^u)}_{\text{self-excitation}}\;+\;\underbrace{\nu_{\text{group}}(t, \Hist^{\text{group}})}_{\text{group-excitation}}\Big),
\end{equation}
where $\phi(\cdot)$ is a non-negative activation (we use softplus) ensuring $\lambda_u(t) > 0$. The three terms target three distinct sources of intensity: $\mu_u$ captures the baseline popularity of chain $u$; $\nu_{\text{self}}$ captures the inertia of $u$'s own history; and $\nu_{\text{group}}$ captures cross-chain mutual excitation. We now realize these terms in turn.

\subsection{Self-excitation via attention (AttNHP)}
\label{subsec:self}

To realize $\nu_{\text{self}}$, we must capture long-range temporal dependencies within the history of a single chain. Following~\citet{yang2022transformer}, we adopt an attention-based encoder that builds a history-aware temporal embedding.

\paragraph{Setup.}
Fix a chain $u = (s, r)$. Suppose we observe $\Gamma - 1$ historical events on $[0, t_\Gamma)$, $\Hist_{t_\Gamma}^u = \{(o_1, t_1), \ldots, (o_{\Gamma-1}, t_{\Gamma-1})\}$ with $0 < t_1 < \cdots < t_{\Gamma-1} < t_\Gamma$. Given a target time $t$ (typically $t = t_\Gamma$), our goal is a temporal embedding $h_u(t) \in \R^D$.

\paragraph{Layer-wise attention updates.}
We stack $L$ attention layers and concatenate the layer-wise representations: $h_u(t) = \Concat\big(h_u^{(0)}(t), h_u^{(1)}(t), \ldots, h_u^{(L)}(t)\big)$. For each layer $\ell > 0$,
\begin{equation}
h_u^{(\ell)}(t) = \underbrace{h_u^{(\ell-1)}(t)}_{\text{residual}} + \tanh\!\left(\frac{\sum_{(o_i, t_i) \in \Hist_t^u} a^{(\ell)}\!\big((o_i, t_i), t\big)\, v^{(\ell)}(o_i, t_i)}{1 + \sum_{(o_j, t_j) \in \Hist_t^u} a^{(\ell)}\!\big((o_j, t_j), t\big)}\right).
\label{eq:attn-update}
\end{equation}
The attention weight between target time $t$ and historical event $(o_i, t_i)$ is
\begin{equation}
a^{(\ell)}\!\big((o_i, t_i), t\big) = \exp\!\left(k^{(\ell)}(o_i, t_i)^\top q^{(\ell)}(t)\,/\,\sqrt{D}\right).
\label{eq:attn-weight}
\end{equation}
Here $q^{(\ell)}(t)$ is the query for the target state, and $k^{(\ell)}(o_i, t_i)$ and $v^{(\ell)}(o_i, t_i)$ are the key and value for the historical event, all produced via learned linear projections $Q^{(\ell)}, K^{(\ell)}, V^{(\ell)} \in \R^{D \times D}$. For any time $\tau \in \{t, t_i\}$,
\begin{align}
q^{(\ell)}(\tau) &= Q^{(\ell)}\big(1;\, [\tau];\, h_u^{(\ell-1)}(\tau)\big), \\
k^{(\ell)}(\tau) &= K^{(\ell)}\big(1;\, [\tau];\, h_u^{(\ell-1)}(\tau)\big), \\
v^{(\ell)}(\tau) &= V^{(\ell)}\big(1;\, [\tau];\, h_u^{(\ell-1)}(\tau)\big),
\end{align}
where $[1; [\tau]; \cdot]$ concatenates a bias term, a time encoding, and the previous-layer representation.

\paragraph{Base case and time encoding.}
As the base case, $h^{(0)}(t) \stackrel{\text{def}}{=} [o]^{(0)}$ is the learned embedding of the object at time $t$. Because continuous time $t \in \R$ cannot be embedded directly, we use sinusoidal time encodings $[t] \in \R^D$:
\begin{equation}
[t]_d = \sin\!\left(\frac{t}{m \cdot \theta^{d/D}}\right) \mathbf{1}(d \text{ even}) + \cos\!\left(\frac{t}{m \cdot \theta^{d/D}}\right) \mathbf{1} (d \text{ odd}),
\label{eq:time-encoding}
\end{equation}
for $0 \leq d < D$, with $m$ and $\theta$ controlling the timescale.

The output of this module is a per-event self-excitation embedding $h_u(t)$ that summarizes the chain's own history. It enters the full intensity in Section~\ref{subsec:intensity}.

\subsection{Group excitation via semantic soft grouping}
\label{subsec:group}

We now realize $\nu_{\text{group}}$. The challenge is to capture mutual excitation across chains \emph{without} exhaustive pairwise computation. Our key idea is threefold: (i) infer a soft group assignment for each chain; (ii) learn group-level Hawkes priors; and (iii) translate those priors into an analytical attention mask, so that a single multi-head attention layer realizes the full Hawkes excitation kernel.

\paragraph{Semantic soft group assignment.}
For chain $u = (s, r)$, we extract static embeddings $e_s$ and $e_r$ of the subject and relation, and map them into a distribution over $G$ latent groups via an MLP with a GELU activation:
\begin{equation}
w_u = \Softmax\!\left(\frac{W_2\, \GELU(W_1 [e_s \,\|\, e_r] + b_1) + b_2}{\tau}\right),
\label{eq:group-assignment}
\end{equation}
where $w_u \in [0, 1]^G$ is the soft assignment, $\|$ denotes concatenation, and $\tau$ is a temperature controlling the sharpness of the distribution. For a mini-batch of $B$ sequences, the stacked weights form $W \in \R^{B \times G}$.

\paragraph{Group-level Hawkes priors.}
We introduce two globally learnable parameters: a raw group-to-group excitation matrix $\Phi \in \R^{G \times G}$ and a group-wise decay vector $\gamma \in \R^G$. Here $\Phi_{m,n}$ controls how strongly an event in group $n$ excites future events in group $m$, and $\gamma_m$ governs the temporal decay rate within group $m$.

Given the soft assignments $W$, we instantiate the \emph{expected} sequence-to-sequence excitation $\bar{\Phi} \in \R^{B \times B}$ and decay $\bar{\gamma} \in \R^{B}$ via
\begin{equation}
\bar{\Phi} = W \Phi W^\top, \qquad \bar{\gamma} = W \gamma.
\label{eq:expected-priors}
\end{equation}
This soft-group construction reduces the cost of cross-sequence excitation from $O(B^2)$ raw parameters to $O(G^2)$, while preserving the ability to express asymmetric, semantically-driven interactions between any two chains in the batch.

\paragraph{Hawkes-derived analytical attention mask.}
Consider a query event indexed by $q$ (occurring at $t_q$ in chain $u$) and a historical key event indexed by $k$ (occurring at $t_k$ in chain $v$), with $\Delta t_{q,k} = t_q - t_k$. We define an event-level attention mask
\begin{equation}
M_{q,k} = \begin{cases}
\log(\bar{\Phi}_{u,v}) - \bar{\gamma}_u\, \Delta t_{q,k}, & \Delta t_{q,k} > 0, \\
-\infty, & \text{otherwise.}
\end{cases}
\label{eq:hawkes-mask}
\end{equation}
This mask is not arbitrary: when the standard multi-head attention applies the exponential during its softmax, the mask collapses exactly into a Group Hawkes intensity kernel,
\begin{equation}
\exp(M_{q,k}) = \bar{\Phi}_{u,v}\, \exp(-\bar{\gamma}_u\, \Delta t_{q,k}).
\end{equation}
Through the exponential inside the softmax, the semantic similarity (the $q^\top k$
term) is \emph{modulated} by the group-Hawkes factor
$\bar{\Phi}_{u,v}\exp(-\bar{\gamma}_u\Delta t)$ and then normalized. The aggregated
$z_u(t)$ thus combines semantic feature matching with a macro-level mutual-excitation
prior.

\paragraph{Cross-chain feature aggregation.}
With the mask defined, we aggregate features across chains. Let $H_{\text{base}} \in \R^{B \times L \times D}$ stack the base context features of the current batch (constructed in Eq.~\eqref{eq:hbase} below by concatenating local temporal features with static semantics). To enable global inter-chain communication, we flatten the keys and values into a shared memory pool $H_{\text{shared}} \in \R^{S \times D}$, where $S = B \times L$. The refined contextual representations are
\begin{equation}
\hat{H} = \LayerNorm\!\Big(\MHA\big(Q = H_{\text{base}},\, K = H_{\text{shared}},\, V = H_{\text{shared}},\, \text{mask} = M + P\big)\Big),
\end{equation}
where $P$ is a Boolean key-padding mask filtering out invalid or padded historical events. Each event is therefore mutually excited and semantically enriched by the relevant history across the entire memory pool, strictly governed by the learned group dynamics.

\subsection{Full conditional intensity}
\label{subsec:intensity}

We now assemble the final intensity. Let $z_u(t)$ denote the refined representation for chain $u$ at time $t$---this is the output of the Hawkes-guided cross-attention from Section~\ref{subsec:group}, applied on top of the self-attention encoder from Section~\ref{subsec:self}. Concretely,
\begin{equation}
z_u(t) = \LayerNorm\!\left(h_{\text{base}}^{(u)}(t) + \sum_{v \in B} \sum_{t_k < t} \alpha_{u,v}(t, t_k)\, v_v(t_k)\right),
\label{eq:zu}
\end{equation}
where the foundational representation explicitly concatenates the micro-level history embedding $h_u(t)$ (from Section~\ref{subsec:self}) with the static semantic embeddings:
\begin{equation}
h_{\text{base}}^{(u)}(t) = \big(h_u(t) \,\|\, e_s \,\|\, e_r\big),
\label{eq:hbase}
\end{equation}
and the cross-attention weight $\alpha_{u,v}$ is \emph{modulated} by the Hawkes mask $M$:
\begin{equation}
\alpha_{u,v}(t, t_k) = \Softmax\!\left(\frac{q_u(t)^\top k_v(t_k)}{\sqrt{D}} + M_{u,v}(t, t_k)\right).
\label{eq:alpha}
\end{equation}
Through the exponential inside the softmax, semantic similarity (the $q^\top k$ term) is multiplied by the physical Hawkes kernel $\bar{\Phi}_{u,v} \exp(-\bar{\gamma}_u \Delta t)$. The aggregated $z_u(t)$ therefore \emph{simultaneously} captures semantic feature matching and macro-level mutual excitation.

To compute the conditional intensity of observing a specific object $o \in \V$, we apply a linear projection followed by softplus to ensure positivity:
\begin{equation}
\lambda_u(o, t) = \Softplus\!\left(w_o^\top z_u(t) + b_o\right),
\label{eq:lambda-final}
\end{equation}
where $w_o$ and $b_o \in \R$ are object-specific parameters: $b_o$ acts as a learned baseline popularity, while $z_u(t)$ provides the dynamic, history-aware drive.

\paragraph{Optimization objective.}
We train by minimizing the negative log-likelihood (NLL) of the observed event streams. For a dataset of chains $\mathcal{U}$, the entity-prediction loss is
\begin{equation}
\Loss_{\text{event}} = -\sum_{u \in \mathcal{U}} \left(\sum_{(o_i, t_i) \in \Hist^u} \log \lambda_u(o_i, t_i) - \int_0^T \sum_{o \in \V} \lambda_u(o, t)\, dt\right).
\label{eq:event-loss}
\end{equation}
We approximate the integral term using the rectangle rule, computed as the product of the time interval and the aggregated intensity at the target time step.

\subsection{Non-Crossing Quantile regression for time prediction}
\label{subsec:ncq}

Equations~\eqref{eq:entity-prediction} and~\eqref{eq:time-prediction} require predicting \emph{when} a future event will occur, not just which entity. We treat this as a distributional regression problem on the inter-arrival time $\Delta_u = t_{\text{next}} - t_{\text{now}}$, using a Non-Crossing Quantile (NCQ) module adapted from~\citet{wu2023dnet}.

\paragraph{Why quantile regression.}
Inter-arrival times in TKGs are heavy-tailed: a small fraction of very long gaps drag the mean far from any typical value. Mean-squared-error (MSE) regression therefore yields biased and unstable point estimates, and provides no uncertainty information. Quantile regression sidesteps both problems by estimating the conditional CDF directly. However, naively predicting a set of quantiles independently can produce \emph{crossings}---e.g., $\hat{\Delta}(0.1) > \hat{\Delta}(0.9)$---that violate CDF monotonicity. NCQ enforces monotonicity by construction.

\paragraph{Architecture: value and delta layers.}
Sharing the same input context $z_u(t)$, we use two independent dense layers---a \emph{Value Layer} $v(\cdot)$ and a \emph{Delta Layer} $d(\cdot)$---that decouple the central tendency from the distributional spread. Given $\Gamma$ sorted target quantiles $\Time = \{\tau_1, \tau_2, \ldots, \tau_\Gamma\}$, the prediction model is
\begin{equation}
\text{NCQ}(z_u(t)) = v(z_u(t); \theta_v) \oplus d(z_u(t); \theta_\delta),
\label{eq:ncq}
\end{equation}
where $\oplus$ denotes broadcast addition, and $\theta_v, \theta_\delta$ are learnable parameters. The Value Layer outputs a scalar $v_t \in \R$ representing the central level of the predicted quantiles. The Delta Layer outputs a vector $c_t \in \R^\Gamma$ that shapes the spread.

\paragraph{Monotonicity via positive increments.}
To guarantee $\hat{\Delta}_u(\tau_1) < \cdots < \hat{\Delta}_u(\tau_\Gamma)$, we must ensure positive gaps between adjacent quantiles. We pass the Delta Layer outputs through softplus,
\begin{equation}
\delta_\gamma = \Softplus([c_t]_\gamma) = \log(1 + e^{[c_t]_\gamma}) > 0, \qquad \gamma = 1, \ldots, \Gamma,
\end{equation}
and construct the $\gamma$-th quantile by adding a centered cumulative sum to the central value:
\begin{equation}
\hat{\Delta}_u(\tau_\gamma) = v_t + \left[\sum_{j=1}^{\gamma} \delta_j - \Gamma^{-1} \sum_{j=1}^{\Gamma} (\Gamma + 1 - j)\, \delta_j\right].
\label{eq:quantile-construction}
\end{equation}
The bracketed term centers the quantiles so that their average equals $v_t$. By construction, $\hat{\Delta}_u(\tau_\gamma) - \hat{\Delta}_u(\tau_{\gamma-1}) = \delta_\gamma > 0$, so monotonicity is guaranteed structurally rather than by penalty.

\paragraph{Optimization.}
The module minimizes the standard average pinball loss,
\begin{equation}
\Loss_{\text{time}} = \Gamma^{-1} \sum_{\gamma=1}^{\Gamma} \rho_{\tau_\gamma}\!\left(\Delta_{\text{true}} - \hat{\Delta}_u(\tau_\gamma)\right),
\label{eq:time-loss}
\end{equation}
where $\rho_\tau(y) = y(\tau - \mathbf{1} (y < 0))$ is the check function. At inference, we use the median quantile ($\tau = 0.5$) as a robust point estimate.

\subsection{Joint training}
\label{subsec:training}

We jointly optimize for \emph{what} and \emph{when} by combining Eqs.~\eqref{eq:event-loss} and~\eqref{eq:time-loss}:
\begin{equation}
\Loss_{\text{total}} = \Loss_{\text{event}} + \beta\, \Loss_{\text{time}},
\label{eq:total-loss}
\end{equation}
where $\beta$ balances the two tasks. Because $\Loss_{\text{time}}$ uses the bounded check function instead of squared error, its gradients remain well-scaled even on heavy-tailed inter-arrival distributions, yielding a stable joint signal. 

We optimize the total objective $\mathcal{L}_{\text{total}}$ using the Adam optimizer with an initial learning rate of $10^{-3}$. Training is performed on mini-batches; each batch consists of $B$ sampled event chains, each containing a sequence of $L$ events. For every event in the sequence, we compute the conditional intensity $\lambda_u(o, t)$ and quantile predictions $\hat{y}_{\tau}$ in a single forward pass. All models are trained for a maximum of 30 epochs, and the model with the minimum validation loss is kept as the final model.

\section{Experiments}
\label{sec:experiments}

We evaluate GAttNHP on three event-based TKG benchmarks in the main paper---\textbf{ICEWS18}~\citep{xu2023cenet}, \textbf{ICEWS14}~\citep{xu2023cenet}, and \textbf{ICEWS05-15}~\citep{liu2022tlogic}. Results on three additional datasets (GDELT, WIKI, YAGO) and comprehensive ablations are deferred to Appendix~\ref{app:additional}. We answer two questions: (Q1) does GAttNHP improve future link prediction over state-of-the-art baselines? (Q2) does the NCQ head deliver more accurate and better-calibrated time prediction than mean-based regression, external TKG baselines, and alternative point-process heads?

\paragraph{Baselines.}
For entity prediction, we compare against three families of methods: (i) \emph{static KG} models---TransE~\citep{bordes2013transe}, DistMult~\citep{yang2014distmult}, ComplEx~\citep{trouillon2016complex}; (ii) \emph{TKG interpolation} models---TTransE~\citep{garcia2018tatranse}, DE-DistMult, DE-SimplE~\citep{goel2020de}, TNTComplEx~\citep{lacroix2020tensor}; and (iii) \emph{TKG extrapolation} models---CyGNet~\citep{zhu2021cygnet}, RE-GCN~\citep{li2021regcn}, CEN~\citep{li2022cen}, CENET~\citep{xu2023cenet}, TLogic~\citep{liu2022tlogic}, GHT~\citep{sun2022ght}, ECEformer~\citep{fang2024eceformer}. For time prediction, we compare our Non-Crossing Quantile (NCQ) head not only against an internal ablative variant of GAttNHP that replaces the NCQ with a standard mean-squared-error (MSE) regressor, but also against state-of-the-art TKG extrapolation models (e.g., GHT~\citep{sun2022ght} and GHNN~\citep{han2020GHNN}) and other widely-adopted continuous-time point process baselines (RQS-QF~\citep{sbt2021}  and RMTPP \citep{du2016recurrent}).

\subsection{Temporal link prediction}
\label{subsec:link-prediction}

Table~\ref{tab:link-prediction} reports time-aware raw MRR and Hits@1/3/10 on the three ICEWS datasets. Across every metric and every dataset, GAttNHP substantially outperforms prior methods. On ICEWS14 in particular, GAttNHP attains an MRR of \textbf{0.5068} ($\pm$ 0.0064), an absolute improvement of \textbf{8.25} points over the previous best, ECEformer (0.4243); its Hits@1 of \textbf{0.4119} ($\pm$ 0.0056) likewise outpaces the runner-up TLogic (0.3186) by nearly ten points. Similar gains hold on ICEWS18 and ICEWS05-15. These results directly validate our two architectural claims: the self-attention encoder captures continuous-time history that snapshot-based baselines miss, and the semantic-soft-grouping mechanism adds cross-chain mutual excitation that no prior TKG model expresses.

\begin{table}[t]
\caption{Temporal link prediction on ICEWS14, ICEWS18, and ICEWS05-15. Metrics are time-aware raw MRR and Hits@1/3/10. \textbf{Bold} marks the best result; the previous SOTA is \underline{underlined}.}
\label{tab:link-prediction}
\centering
\small
\setlength{\tabcolsep}{3pt}
\makebox[\textwidth][c]{
\begin{tabular}{l cccc cccc cccc}
\toprule
& \multicolumn{4}{c}{ICEWS18} & \multicolumn{4}{c}{ICEWS14} & \multicolumn{4}{c}{ICEWS05-15} \\
\cmidrule(lr){2-5} \cmidrule(lr){6-9} \cmidrule(lr){10-13}
Method & H@1 & H@3 & H@10 & MRR & H@1 & H@3 & H@10 & MRR & H@1 & H@3 & H@10 & MRR \\
\midrule
TransE      & 0.0482 & 0.1468 & 0.2896 & 0.1273 & 0.0162 & 0.2718 & 0.4604 & 0.1737 & 0.0545 & 0.2797 & 0.4573 & 0.1961 \\
DistMult    & 0.0292 & 0.0691 & 0.1501 & 0.0704 & 0.0256 & 0.0711 & 0.1571 & 0.0695 & 0.0862 & 0.1731 & 0.2953 & 0.1569 \\
ComplEx     & 0.1016 & 0.2623 & 0.4111 & 0.2100 & 0.1519 & 0.3453 & 0.4849 & 0.2732 & 0.1529 & 0.3779 & 0.5400 & 0.2924 \\
TTransE     & 0.0164 & 0.0777 & 0.1969 & 0.0769 & 0.0228 & 0.1534 & 0.3399 & 0.1259 & 0.0283 & 0.1619 & 0.3444 & 0.1313 \\
DE-DistMult & 0.1352 & 0.2518 & 0.4066 & 0.2247 & 0.2262 & 0.3475 & 0.4835 & 0.3128 & 0.2389 & 0.3792 & 0.5321 & 0.3378 \\
DE-SimplE   & 0.1401 & 0.2669 & 0.4226 & 0.2341 & 0.2376 & 0.3779 & 0.5050 & 0.3304 & 0.2491 & 0.3940 & 0.5431 & 0.3495 \\
TNTComplEx  & 0.0763 & 0.2158 & 0.3559 & 0.1741 & 0.1374 & 0.3199 & 0.4603 & 0.2546 & 0.1571 & 0.3792 & 0.5400 & 0.2950 \\
RE-GCN      & 0.1663 & 0.2938 & 0.4375 & 0.2582 & 0.2546 & 0.3876 & 0.5337 & 0.3473 & 0.2607 & 0.4109 & 0.5674 & 0.3641 \\
CyGNet      & 0.1302 & 0.2381 & 0.3601 & 0.2083 & 0.2557 & 0.3825 & 0.5008 & 0.3423 & 0.2597 & 0.4091 & 0.5472 & 0.3598 \\
CEN         & 0.1353 & 0.2438 & 0.3800 & 0.2176 & 0.2333 & 0.3534 & 0.4852 & 0.3192 & 0.2581 & 0.3996 & 0.5461 & 0.3562 \\
CENET       & 0.1775 & 0.3138 & 0.4656 & 0.2750 & 0.2794 & 0.4252 & 0.5733 & 0.3781 & 0.2957 & 0.4566 & 0.6021 & 0.4017 \\
TLogic      & \underline{0.1869} & \underline{0.3259} & \underline{0.4784} & \underline{0.2832} & \underline{0.3186} & 0.4754 & 0.6096 & 0.4187 & \underline{0.3447} & \underline{0.5251} & \underline{0.6729} & \underline{0.4586} \\
GHT         & 0.1808 & 0.3076 & 0.4576 & 0.2740 & 0.2777 & 0.4166 & 0.5619 & 0.3740 & 0.3079 & 0.4685 & 0.6273 & 0.4150 \\
ECEformer   & 0.1431 & 0.2623 & 0.4203 & 0.2339 & 0.3084 & \underline{0.4796} & \underline{0.6505} & \underline{0.4243} & 0.3014 & 0.4728 & 0.6379 & 0.4166 \\
\midrule
\multirow{2}{*}{\textbf{GAttNHP}} 
& \textbf{0.2825} & \textbf{0.4414} & \textbf{0.5821} & \textbf{0.3863} & \textbf{0.4119} & \textbf{0.5637} & \textbf{0.6775} & \textbf{0.5068} & \textbf{0.4101} & \textbf{0.5786} & \textbf{0.6996} & \textbf{0.5137} \\
& {\scriptsize $\pm$ 0.0156} & {\scriptsize $\pm$ 0.0217} & {\scriptsize $\pm$ 0.0234} & {\scriptsize $\pm$ 0.0184} & {\scriptsize $\pm$ 0.0056} & {\scriptsize $\pm$ 0.0094} & {\scriptsize $\pm$ 0.0073} & {\scriptsize $\pm$ 0.0064} & {\scriptsize $\pm$ 0.0071} & {\scriptsize $\pm$ 0.0099} & {\scriptsize $\pm$ 0.0057} & {\scriptsize $\pm$ 0.0072} \\
\bottomrule
\end{tabular}}
\end{table}

\subsection{Time prediction}
\label{subsec:time-prediction}

Time-prediction evaluation proceeds along two complementary axes.
\textbf{External TKG baselines.} We first compare GAttNHP-NCQ against two external TKG
occurrence-time methods, GHT~\citep{sun2022ght} and GHNN~\citep{han2020GHNN}, together with an internal mean-based variant (GAttNHP-MSE). As shown in Table~\ref{tab:time-prediction}
(MAE in days; we use the median $\tau=0.5$ as the point estimate), GAttNHP-NCQ attains the lowest MAE on all three benchmarks, improving over GAttNHP-MSE by $47\%$/$50\%$/$70\%$ and over the stronger external baseline GHT by $55\%$/$73\%$/$29\%$ on ICEWS14/ICEWS18/ICEWS05-15. The advantage over MSE widens on the more volatile datasets:
heavy-tailed inter-arrival times amplify the squared-error penalty into very large gradients on outliers that destabilize training, whereas NCQ's bounded pinball loss avoids this pathology.
\textbf{Decoder-head comparison.} To isolate the contribution of the NCQ head itself, we keep the GAttNHP encoder fixed and replace \emph{only} the time-prediction decoder, comparing NCQ against the rational-quadratic-spline quantile head RQS-QF~\citep{sbt2021} and
the RMTPP head~\citep{du2016recurrent}. Since all variants share the identical embedding learning, data, and protocol, any difference reflects the decoder head alone.
Table~\ref{tab:time-head-compare} shows that NCQ again attains the lowest MAE and the best-calibrated intervals across datasets: its empirical $90\%$-interval coverage stays close to nominal ($0.85$--$0.90$), whereas RQS-QF is severely under-covered ($0.21$--$0.58$) with far larger interval scores, and RMTPP lies in between. We further emphasize that the MSE variant produces no distributional output at all, underscoring the value of the quantile formulation.

\begin{table}[t]
\caption{Occurrence-time prediction (MAE in days, lower is better). \textbf{Bold}: best.}
\label{tab:time-prediction}
\centering
\begin{tabular}{lcccc}
\toprule
Dataset    & GAttNHP-MSE & GHT  & GHNN & GAttNHP-NCQ (Ours) \\
\midrule
ICEWS14    & 2.52  & 2.95 & 5.80 & \textbf{1.33} \\
ICEWS18    & 1.67  & 3.07 & 4.45 & \textbf{0.83} \\
ICEWS05-15 & 10.33 & 4.32 & 6.93 & \textbf{3.07} \\
\bottomrule
\end{tabular}
\end{table}
 
\begin{table}[t]
\centering
\small 
\setlength{\tabcolsep}{4pt} 
\caption{Time-prediction head comparison on the same GAttNHP encoder
(NCQ is ours). QSm $=$ mean pinball; Cov@$p$ targets the
nominal level $p$. Lower is better for all metrics except coverage, where \textbf{bold}
marks the value closest to the nominal level; for the remaining columns \textbf{bold}
marks the best (lowest). ``Time (s)'' is the average training time per epoch.}
\label{tab:time-head-compare}
\begin{tabular}{l l cccc cc cc c}
\toprule
\multirow{2}{*}{Dataset} & \multirow{2}{*}{Head Variant} & \multirow{2}{*}{SMAPE} & \multicolumn{3}{c}{Quantile Score (QS)} & \multicolumn{2}{c}{Coverage} & \multicolumn{2}{c}{Interval Score} & \multirow{2}{*}{Time (s)} \\
\cmidrule(lr){4-6} \cmidrule(lr){7-8} \cmidrule(lr){9-10}
& & & QSm & QS50 & QS95 & Cov@50 & Cov@90 & IS50 & IS90 & \\
\midrule
\multirow{3}{*}{ICEWS14}
 & RQS   & 0.97 & 2.74 & 2.88 & 4.88 & 0.17 & 0.29 & 11.19 & 52.10 & 22.60 \\
 & RMTPP & 0.69 & 1.06 & 1.51 & 0.77 & 0.61 & 0.79 &  5.37 & 10.97 & 43.94 \\
 & \textbf{NCQ (Ours)} & \textbf{0.69} & \textbf{0.82} & \textbf{1.33} & \textbf{0.44} & \textbf{0.46} & \textbf{0.85} & \textbf{4.13} & \textbf{7.32} & \textbf{7.23} \\
\midrule
\multirow{3}{*}{ICEWS18}
 & RQS   & 0.92 & 2.14 & 2.22 & 3.89 & 0.429 & 0.58 &  8.72 & 41.40 & 104.46 \\
 & RMTPP & 1.01 & 1.09 & 1.40 & 1.01 & \textbf{0.50} & 0.63 &  5.51 & 12.83 & 185.13 \\
 & \textbf{NCQ (Ours)} & \textbf{0.89} & \textbf{0.55} & \textbf{0.83} & \textbf{0.35} & 0.55 & \textbf{0.90} & \textbf{2.73} & \textbf{5.50} & \textbf{39.40} \\
\midrule
\multirow{3}{*}{ICEWS05-15}
 & RQS   & 1.19 & 21.95 & 22.01 & 41.56 & 0.14 & 0.21 & 87.92 & 437.82 & 82.76 \\
 & RMTPP & 0.64 &  6.00 &  7.89 &  4.75 & 0.67 & 0.88 & 30.41 &  68.83 & 136.79 \\
 & \textbf{NCQ (Ours)} & \textbf{0.47} & \textbf{1.99} & \textbf{3.07} & \textbf{1.12} & \textbf{0.55} & \textbf{0.90} & \textbf{9.93} & \textbf{19.26} & \textbf{37.03} \\
\bottomrule
\end{tabular}
\end{table}

\section{Conclusion}
\label{sec:conclusion}

We introduced \textbf{GAttNHP}, a unified framework for extrapolative reasoning on temporal knowledge graphs that bridges structural representation learning with continuous-time point processes. Built on a self-attention neural Hawkes encoder, GAttNHP captures long-range dependencies within each event chain. Its central contribution is the macro-level group interaction module, which---through a semantic soft-grouping strategy that translates globally learnable Hawkes priors into an analytical attention mask---lets event chains share excitation patterns through their latent group memberships rather than through pairwise interaction. A Non-Crossing Quantile head completes the picture, providing calibrated, monotonically ordered time predictions that remain stable under heavy-tailed inter-arrival distributions and, as a side benefit, supply cleaner gradients for the joint objective. Across six benchmarks, GAttNHP achieves state-of-the-art results, with its largest improvements concentrated on the long-tail event chains where existing TKG methods fail most. Future work will explore adaptive group structures and extend the point-process formulation to multi-hop reasoning.

\begin{ack}
We thank the anonymous referees and the meta-reviewer
for their constructive comments, which have significantly
improved this manuscript. Tang and Tian's research was
supported by the National Key R\&D Program of China (Grant No. 2022YFA1003701).
\end{ack}


\appendix

\section{Appendix}
\label{app:additional}

\noindent\emph{The appendix is preserved from the original manuscript: Datasets and Evaluation (dataset statistics, evaluation metrics), experimental Setup (Implementation Details, Hyperparameter Search and Tuning Budget, Baseline Reproduction, Efficiency and scalability), additional results on GDELT/WIKI/YAGO, ablation studies (component analysis, sensitivity to the number of latent groups $G$), the frequency-aware analysis demonstrating that the group-interaction module's largest gains arise on long-tail event chains, and the qualitative case study on ICEWS14.} 

\subsection{Datasets and Evaluation}

\subsubsection{Dataset Statistics}
We evaluate our proposed method on six benchmark datasets, including four event-centric datasets (ICEWS14, ICEWS05-15, ICEWS18, GDELT) and two knowledge-centric datasets (WIKI, YAGO). The detailed statistics for each dataset are summarized in Table~\ref{tab:stats}.

\begin{table}[ht]
	\centering
	\caption{Statistics of the benchmark datasets.}
	\label{tab:stats}
	\resizebox{\columnwidth}{!}{
		\begin{tabular}{lccccccc}
			\toprule
			Dataset & Ent. & Rel. & Time & Interval & Train & Valid & Test \\
			\midrule
			ICEWS14    & 7,128  & 230 & 365   & 24 hours & 74,845    & 8,514   & 7,371 \\
			ICEWS05-15 & 10,488 & 251 & 4,017 & 24 hours & 38,692    & 46,092  & 46,275 \\
			ICEWS18    & 23,033 & 256 & 7,272 & 1 hours & 373,018   & 45,995  & 49,545 \\
			GDELT      & 7,691  & 240 & 8,925 & 15 mins  & 1,734,399 & 238,765 & 305,241 \\
			WIKI       & 12,554 & 24  & 189   & 1 year   & 539,286   & 67,538  & 63,110 \\
			YAGO       & 10,623 & 10  & 232   & 1 year   & 161,540   & 19,523  & 20,026 \\
			\bottomrule
		\end{tabular}
	}
\end{table}

\subsubsection{Evaluation Metrics}
\label{app:metrics}
We evaluate over all test events $i=1,\dots,N$. For entity prediction, $\mathrm{rank}_i$ is
the rank of the ground-truth object; for time prediction, $\Delta_i$ is the true
inter-arrival time and $q_i(\alpha)$ the predicted $\alpha$-quantile, with the median
$q_i(0.5)$ as the point estimate.

\paragraph{Entity Prediction Metrics.}
We evaluate entity prediction performance using two standard metrics: \textbf{Mean Reciprocal Rank (MRR)} and \textbf{Hits@k}. 
Let $\mathcal{S}$ denote the set of query samples (e.g., the test set). For the $j$-th sample, let $\mathcal{Z}_j$ represent the set of ground-truth entities, and $|\mathcal{Z}_j|$ be the number of true labels. We denote the rank of a candidate entity $q$ in the prediction list as $\text{rank}_q$.

\textbf{Hits@k} measures the proportion of instances where the true entity appears among the top $k$ ranked candidates:
\begin{equation}
\text{Hits@k} = \frac{1}{\sum_{j \in \mathcal{S}} |\mathcal{Z}_j|} \sum_{j \in \mathcal{S}} \sum_{q \in \mathcal{Z}_j} \mathbb{I}(\text{rank}_q \leq k),
\end{equation}
where $\mathbb{I}(\cdot)$ is the indicator function.

\textbf{Mean Reciprocal Rank (MRR)} represents the average of the reciprocal ranks of the inferred true entities:
\begin{equation}
\text{MRR} = \frac{1}{\sum_{j \in \mathcal{S}} |\mathcal{Z}_j|} \sum_{j \in \mathcal{S}} \sum_{q \in \mathcal{Z}_j} \frac{1}{\text{rank}_q}.
\end{equation}

\paragraph{Time prediction metrics.}
For point accuracy we use the Mean Absolute Error and the Symmetric MAPE:
\begin{equation}
\mathrm{MAE}=\frac{1}{N}\sum_{i=1}^{N}\big|\Delta_i-q_i(0.5)\big|,
\qquad
\mathrm{SMAPE}=\frac{1}{N}\sum_{i=1}^{N}
\frac{2\,|\Delta_i-q_i(0.5)|}{|\Delta_i|+|q_i(0.5)|}.
\end{equation}
To assess the full predictive distribution we additionally report quantile-based scores.
The quantile (pinball) score at level $\alpha$ and its mean over the evaluated levels
$\mathcal{A}=\{0.05,0.25,0.5,0.75,0.95\}$ are
\begin{equation}
\mathrm{QS}(\alpha)=\frac{1}{N}\sum_{i=1}^{N}
2\big(\mathbf{1}\{\Delta_i\le q_i(\alpha)\}-\alpha\big)\big(q_i(\alpha)-\Delta_i\big),
\qquad
\mathrm{QSm}=\frac{1}{|\mathcal{A}|}\sum_{\alpha\in\mathcal{A}}\mathrm{QS}(\alpha).
\end{equation}
We report $\mathrm{QS}(0.95)$ as a tail score; note that $\mathrm{QS}(0.5)$ coincides with the
MAE and is therefore omitted. Calibration is measured by the Mean Absolute Calibration Error
\begin{equation}
\mathrm{MACE}=\frac{1}{|\mathcal{A}|}\sum_{\alpha\in\mathcal{A}}
\Big|\,\alpha-\frac{1}{N}\sum_{i=1}^{N}\mathbf{1}\{\Delta_i\le q_i(\alpha)\}\,\Big|.
\end{equation}
For a central $(1-\rho)\times 100\%$ prediction interval $[\,l_i,u_i\,]=[\,q_i(\rho/2),\,q_i(1-\rho/2)\,]$,
we report the empirical coverage and the interval score
\begin{align}
\mathrm{Cov@}p &=\frac{1}{N}\sum_{i=1}^{N}\mathbf{1}\{\,l_i\le \Delta_i\le u_i\,\},
\qquad p=(1-\rho)\times 100\%, \\
\mathrm{IS}   &=\frac{1}{N}\sum_{i=1}^{N}\Big[(u_i-l_i)
+\tfrac{2}{\rho}(l_i-\Delta_i)\mathbf{1}\{\Delta_i<l_i\}
+\tfrac{2}{\rho}(\Delta_i-u_i)\mathbf{1}\{\Delta_i>u_i\}\Big].
\end{align}
We use the $50\%$ interval ($\rho=0.5$, i.e.\ $[q_i(0.25),q_i(0.75)]$) and the $90\%$ interval
($\rho=0.1$, i.e.\ $[q_i(0.05),q_i(0.95)]$), reported as Cov@50/Cov@90 and IS50/IS90.
For all metrics except coverage, lower is better; for coverage, values close to the nominal
level ($0.50$, $0.90\%$) are better.

\subsection{Experimental Setup}
\label{Experimental Setup}
\subsubsection{Implementation Details}
All models are implemented in PyTorch and optimized with Adam at a learning rate of
$10^{-3}$. Each dataset is split chronologically into training / validation / test sets
($80\%/10\%/10\%$). Unless otherwise stated, GAttNHP uses a hidden size of $64$, a
$16$-dimensional time embedding, $2$ attention layers with $4$ heads, dropout $0.1$,
$G=4$ latent groups, a batch size of $16$, and is trained for $30$ epochs. The time-loss
weight $\beta$ is set to $0.05$ for the denser ICEWS14, ICEWS05-15, ICEWS18,WIKI and YAGO, and to $0.001$ for GDELT. We keep the checkpoint with the lowest training
loss and report test-set metrics. For reproducibility, we fix all random seeds and enable
deterministic CuDNN / algorithm settings. All GAttNHP experiments run on a single NVIDIA
RTX~4080 (16\,GB); the largest baseline (ECEformer) requires a 48\,GB GPU.

\subsubsection{Hyperparameter Search and Tuning Budget}
We grid-search the batch size over $\{4,8,16,32\}$, the maximum number of epochs over
$\{30,50\}$, the number of latent groups $G$ over $\{1,2,4,8,16\}$, and the time-loss
weight $\beta$ over $\{1,0.5,0.1,0.05,0.01,0.001\}$; the remaining architectural settings
(learning rate, hidden size, time-embedding size, number of layers and heads, dropout) are
fixed. Every configuration is selected by validation performance. The selected setting
($G=4$, batch size $16$, $30$ epochs) is used in all main experiments.

\subsubsection{Baseline Reproduction}
For a fair comparison, results for the entity-prediction baselines
[TransE~\citep{bordes2013transe}, DistMult~\citep{yang2014distmult}, TTransE~\citep{garcia2018tatranse}\footnote{\url{https://github.com/LiaoMengqi/KGMH}}, DE-DistMult, DESimplE~\citep{goel2020de}\footnote{\url{https://github.com/BorealisAI/de-simple}},ComplEx~\citep{trouillon2016complex},TNTComplEx~\citep{lacroix2020tensor}\footnote{\url{https://github.com/NacyNiko/TNTComplEx}}; CyGNet~\citep{zhu2021cygnet}\footnote{\url{https://github.com/CunchaoZ/CyGNet}}, RE-GCN~\citep{li2021regcn}\footnote{\url{https://github.com/Lee-zix/RE-GCN}}, CEN~\citep{li2022cen}\footnote{\url{https://github.com/Lee-zix/CEN}}, CENET~\citep{xu2023cenet}\footnote{\url{https://github.com/xyjigsaw/CENET}}, TLogic~\citep{liu2022tlogic}\footnote{\url{https://github.com/liu-yushan/tlogic}}, GHT~\citep{sun2022ght}\footnote{\url{https://github.com/GSYfate/GHT}}, ECEformer~\citep{fang2024eceformer}\footnote{\url{https://github.com/seeyourmind/TKGElib}}] are re-run with their official code on our chronological splits. For occurrence-time prediction, the external baselines GHT~\citep{sun2022ght} and GHNN~\citep{han2020GHNN}\footnote{\url{https://github.com/Jeff20100601/GHNN_clean}} are re-run; GAttNHP-MSE is an internal variant that shares our encoder and only replaces the NCQ head with a mean-squared-error regressor. The additional time-prediction heads (RQSQF~\citep{sbt2021}\footnote{\url{https://github.com/bsouhaib/qf-tpp}}, RMTPP, Exponential) are integrated on top of the \emph{identical} GAttNHP encoder and trained with their own native objectives, so that observed differences reflect the time-prediction head alone.

\paragraph{Efficiency and scalability.}
Table~\ref{tab:cost} reports the trainable parameter count, peak GPU memory, and per-epoch training / inference time of GAttNHP across all six benchmarks, together with the parameter count of the strongest baseline, ECEformer. GAttNHP is markedly lightweight:it uses $2.8$--$8.5$M parameters, i.e.\ roughly $12$--$35\times$ fewer than ECEformer's $88$--$107$M, and its peak memory stays at $2.5$--$3.1$\,GB throughout, so every experiment fits on a single 16\,GB GPU (RTX~4080), whereas ECEformer requires a 48\,GB GPU. Notably,
the memory footprint remains essentially flat even on GDELT (over $1.7$M training facts), because cross-chain excitation is pooled within each mini-batch rather than over the entire graph. Training time scales gracefully with dataset size, from $6.8$\,s/epoch on ICEWS14 to $172$\,s/epoch on the much larger GDELT, with inference taking a few seconds at most. These savings are a direct consequence of the $\mathcal{O}(G^2)$ semantic soft-grouping, which replaces the $\mathcal{O}(B^2)$/$\mathcal{O}(N^2)$ pairwise excitation that a naive cross-chain Hawkes model would require---making GAttNHP both accurate and practical to train under modest hardware.

\begin{table}[t]
\centering
\caption{Parameter count, peak GPU memory, and per-epoch training / inference time.
GAttNHP runs on a single 16\,GB GPU (RTX~4080); ECEformer requires a 48\,GB GPU.}
\label{tab:cost}
\small
\begin{tabular}{lccccc}
\toprule
Dataset & \#Params & \#Params (ECEformer) & Peak Mem (GB) & Train/epoch (s) & Infer (s) \\
\midrule
ICEWS14    & 2.8M & 98.6M  & 2.54 & 6.84   & 0.44  \\
ICEWS18    & 8.5M & 106.6M & 2.87 & 38.97  & 2.88  \\
ICEWS05-15 & 4.0M & --     & 2.61 & 37.56  & 2.05  \\
GDELT      & 3.0M & 88.4M  & 3.12 & 172.40 & 21.17 \\
WIKI       & 4.7M & --     & 2.63 & 17.53  & 2.26  \\
YAGO       & 4.1M & --     & 2.59 & 8.70   & 1.09  \\
\bottomrule
\end{tabular}
\end{table}

\subsection{Additional Results}
To comprehensively evaluate the generalization and robustness of our proposed method, we present additional temporal link prediction results across three widely-used benchmark datasets: GDELT, WIKI, and YAGO. We compare GAttNHP against a diverse set of strong baselines, ranging from traditional static knowledge graph embeddings to recent state-of-the-art continuous-time point process models. The overall evaluation results are summarized in Table \ref{tab:temporal-link-pred-2}.
\begin{table*}[ht]
    \centering
    \caption{Experimental results of temporal link prediction on GDELT, WIKI, and YAGO. Evaluation metrics are time-aware raw MRR and Hits@1/3/10. The best results are boldfaced, and the results of previous SOTAs are underlined.}
    \label{tab:temporal-link-pred-2}
    \resizebox{\textwidth}{!}{%
       \begin{tabular}{lcccccccccccc}
        \toprule
        \multirow{2}{*}{Method} & \multicolumn{4}{c}{GDELT} & \multicolumn{4}{c}{WIKI} & \multicolumn{4}{c}{YAGO} \\
        \cmidrule(lr){2-5} \cmidrule(lr){6-9} \cmidrule(lr){10-13} 
        & h@1 & h@3 & h@10 & MRR & h@1 & h@3 & h@10 & MRR & h@1 & h@3 & h@10 & MRR  \\
        \midrule
        TransE & 0.0000 & 0.1129 & 0.2601 & 0.0897 & 0.2169 & 0.3931 & 0.4668 & 0.3141 & 0.1485 & 0.5014 & 0.5894 & 0.3347 \\
        DistMult & 0.0770 & 0.1249 & 0.2228 & 0.1273 & 0.1176 & 0.2101 & 0.3125 & 0.1828 & 0.0522 & 0.1388 & 0.2706 & 0.1229 \\
        ComplEx & 0.0699 & 0.1864 & 0.3203 & 0.1584 & 0.1379 & 0.3267 & 0.4066 & 0.2440 & 0.2397 & 0.5116 & 0.6254 & 0.3901 \\
        TTransE  & 0.0014 & 0.0074 & 0.0274 & 0.0144 & 0.0686 & 0.1577 & 0.2575 & 0.1333 & 0.1005 & 0.2702 & 0.4004 & 0.2089 \\
        DE-DistMult  & 0.1102 & 0.1857 & 0.3019 & 0.1765 & 0.2331 & 0.3277 & 0.3993 & 0.2945 & 0.3879 & 0.5300 & 0.6180 & 0.4724 \\
        DE-SimplE   & 0.1138 & 0.1933 & 0.3156 & 0.1829 & 0.2328 & 0.3294 & 0.3993 & 0.2945 & 0.3910 & 0.5334 & 0.6243 & 0.4762 \\
        TNTComplEx   & 0.0681 & 0.1861 & 0.3209 & 0.1572 & 0.1370 & 0.3218 & 0.4021 & 0.2421 & 0.2194 & 0.4668 & 0.5657 & 0.3568 \\
        RE-GCN & 0.1164 & 0.1988 & 0.3218 & 0.1865 & 0.2521 & 0.3429 & 0.4157 & 0.3131 & 0.3843 & 0.5304 & 0.6231 & 0.4718 \\
        CyGNet & 0.0896 & 0.1242 & 0.1912 & 0.1267 & \underline{0.3853} & \underline{0.4876} & 0.5370 & \underline{0.4456} & \underline{0.3975} & 0.5805 & 0.6536 & 0.4987 \\
        CEN  & 0.1098 & 0.1867 & 0.3043 & 0.1768 & 0.2316 & 0.3150 & 0.3860 & 0.2880 & 0.3230 & 0.4274 & 0.4972 & 0.3878 \\
        CENET & 0.1003 & 0.1801 & 0.2964 & 0.1685 & 0.3168 & 0.4542 & \underline{0.5617} & 0.4057 & \textbf{0.4873} & \underline{0.6925} & \textbf{0.8126} & \textbf{0.6060} \\
        Tlogic & - & - & - & - & - & - & - & - & 0.4539 & \textbf{0.7031} & \underline{0.7822} & \underline{0.5831} \\
        GHT & \underline{0.1268} & \underline{0.2137} & \underline{0.3442} & 0.2004 & - & - & - & - & - & - & - & - \\
        Eceformer & 0.1158 & 0.2191 & 0.3703 & \underline{0.2018} & 0.1156 & 0.2256 & 0.3963 & 0.2071 & 0.0189 & 0.0648 & 0.1381 & 0.0631 \\
        \midrule
        \textbf{GAttNHP} & \textbf{0.1715} & \textbf{0.2998} & \textbf{0.4398} & \textbf{0.2643} & \textbf{0.4445} & \textbf{0.5619} & \textbf{0.6226} & \textbf{0.5128} & 0.4017 & 0.6058 & 0.6921 & 0.5130 \\
        \bottomrule
    \end{tabular}%
    }
\end{table*}

\paragraph{Performance on Large-Scale and Noisy Data (GDELT \& WIKI)}
GAttNHP exhibits exceptional robustness on large-scale and noisy datasets. On \textit{GDELT}, it achieves an MRR of \textbf{0.2643}, outperforming the second-best method, Eceformer, by a remarkable margin of \textbf{6.25 percentage points}. Likewise, on \textit{WIKI}, GAttNHP establishes a new state-of-the-art performance with an MRR of \textbf{0.5128} and a Hits@1 of \textbf{0.4445}. These results strongly highlight its superior capability in capturing both complex temporal dynamics and topological evolution within densely interacting temporal knowledge graphs.

\paragraph{Performance on Knowledge-Centric Data (YAGO)} 
On the \textit{YAGO} dataset, GAttNHP remains highly competitive, achieving an MRR of \textbf{0.5130}. While it significantly outperforms strong baselines like RE-GCN and CyGNet, it trails slightly behind CENET. We attribute this performance gap to the intrinsic nature of the YAGO dataset: it is dominated by long-term, relatively stable ``facts'' (e.g., entity attributes) rather than instantaneous, event-like interactions. Since our Hawkes-process-based architecture is fundamentally designed to capture mutual excitation effects and bursty dynamics, it naturally yields the most substantial gains on event-driven datasets (e.g., ICEWS and GDELT), while offering a more modest advantage on static-fact-driven benchmarks. Nonetheless, the overall empirical results consistently demonstrate that explicitly modeling group-wise intensity and addressing quantile crossing can effectively improve predictive performance across diverse scenarios.

\subsection{Ablation Study}\label{sec:ablation}

\subsubsection{Analysis of Model Components}

\begin{figure*}[ht]
  \centering

  \begin{minipage}{0.6\linewidth}
    \centering
    \includegraphics[width=\linewidth]{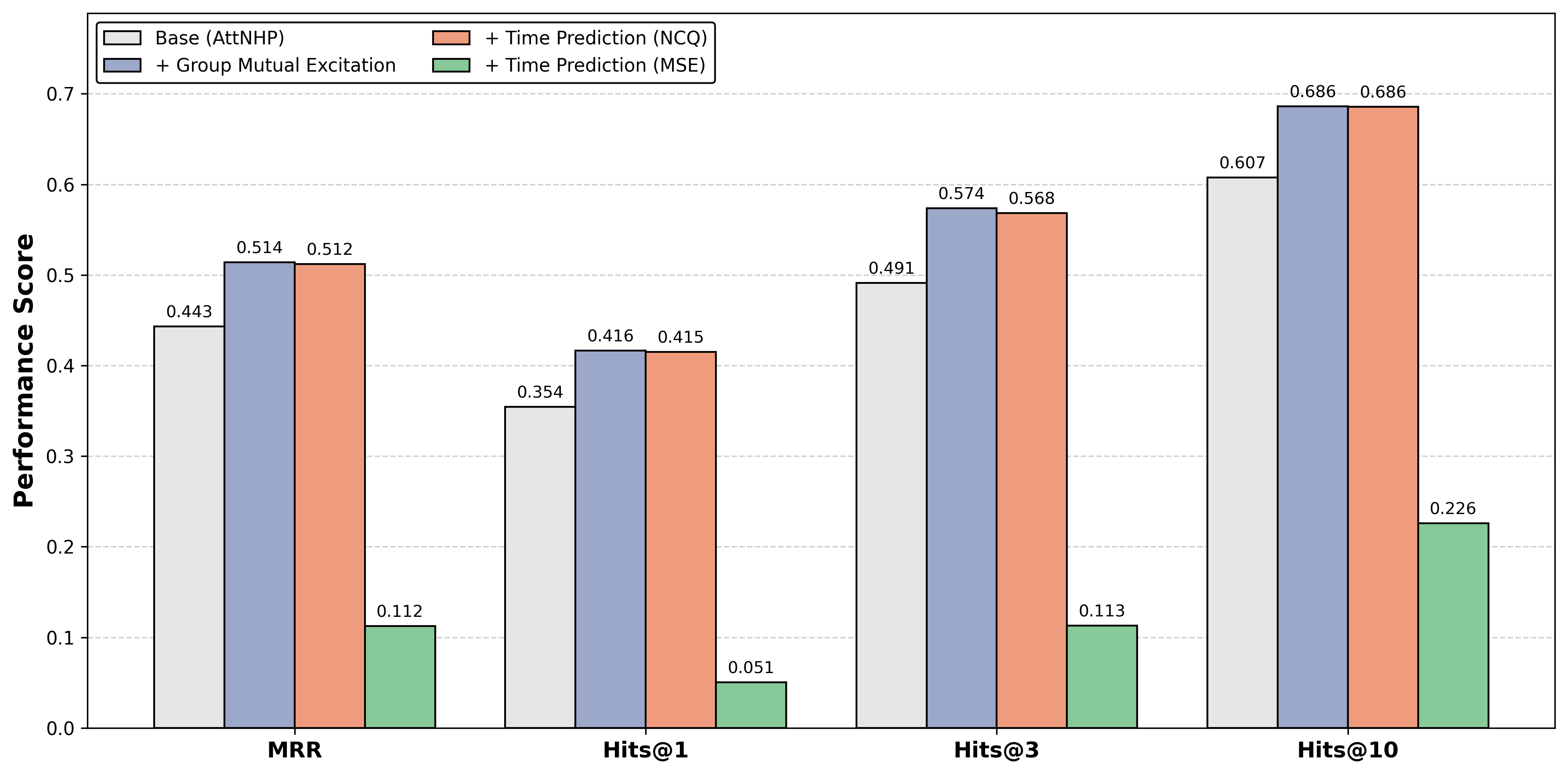}\\[2pt]
    \small (a) ICEWS14
  \end{minipage}
  
  \vspace{1em}
  
  \begin{minipage}{0.6\linewidth}
    \centering
    \includegraphics[width=\linewidth]{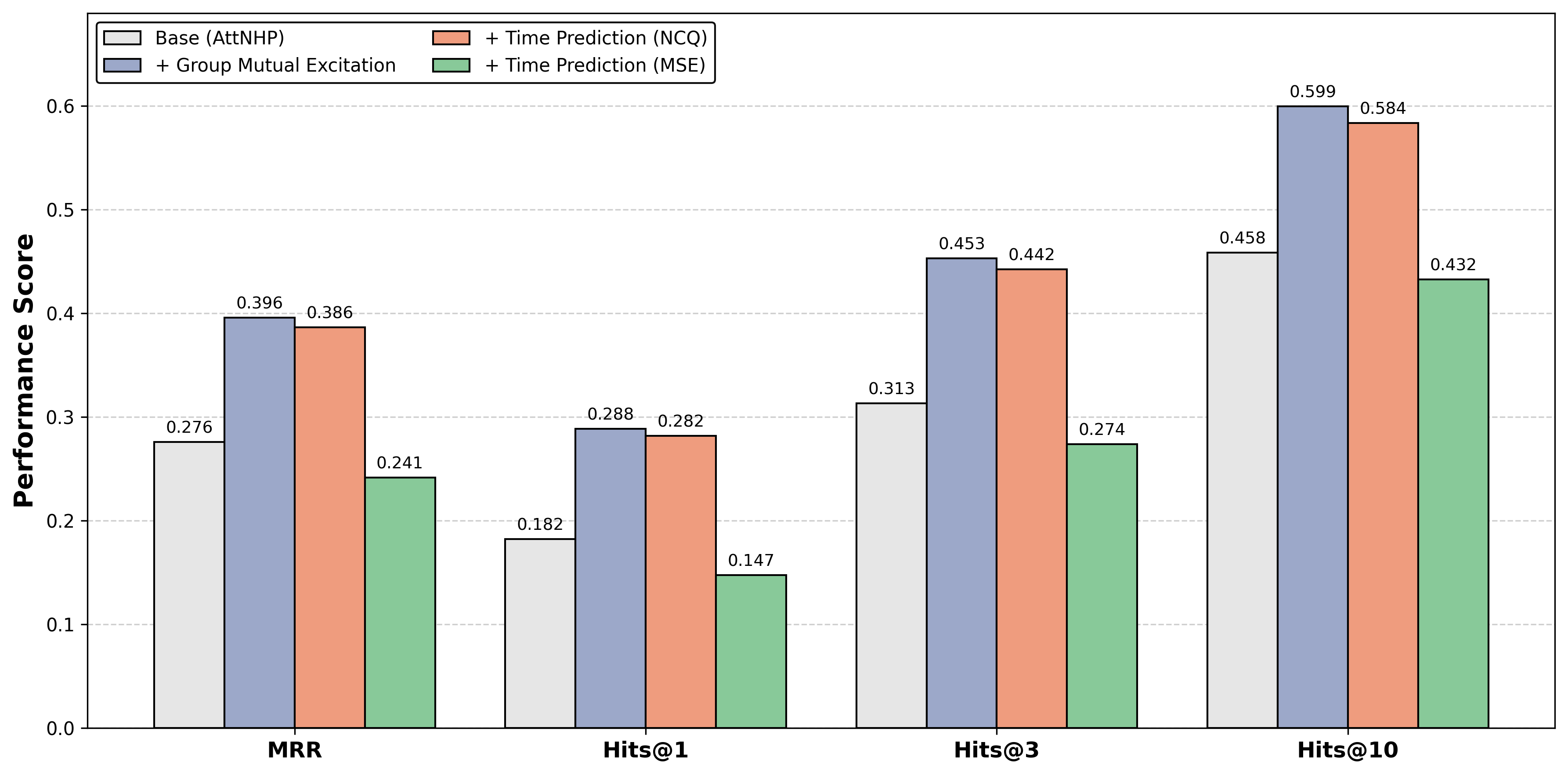}\\[2pt]
    \small (b) ICEWS18
  \end{minipage}
  
  \vspace{1em}
  
  \begin{minipage}{0.6\linewidth}
    \centering
    \includegraphics[width=\linewidth]{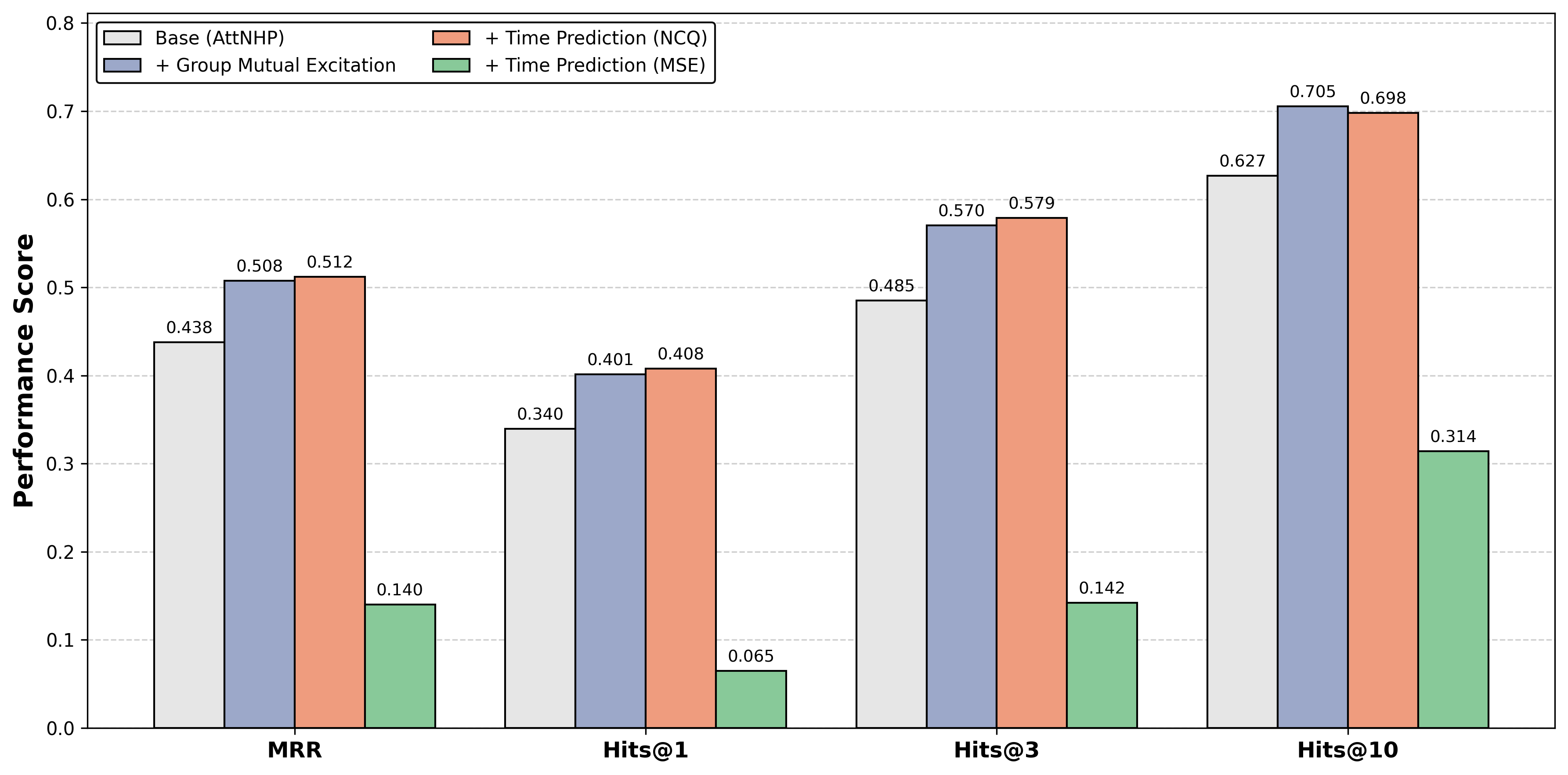}\\[2pt]
    \small (c) ICEWS05-15
  \end{minipage}

  \caption{Ablation study results across ICEWS14, ICEWS18, and ICEWS05-15 datasets. We compare the Base model (Self-Attn Only) with the inclusion of Group Mutual Excitation, and further analyze the Full Model using Non-Crossing Quantile (NCQ) regression versus Mean Squared Error (MSE) for time prediction.}
  \label{fig:ablation_study}
\end{figure*}

To rigorously quantify the contribution of each component in our framework, we conduct an ablation study on \textit{ICEWS14}, \textit{ICEWS18}, and \textit{ICEWS05-15}. We construct four model variants to isolate (i) the effect of the self-attention temporal encoder, (ii) the benefit of the proposed group-wise mutual excitation mechanism, and (iii) the impact of alternative time-prediction objectives. The results are summarized in Figure~\ref{fig:ablation_study}.

The model variants are defined as follows:

    (i) \textbf{Base (Self-Attn Only):} This variant uses only the self-attention temporal encoder to model self-excitation from an event chain’s own history. It ignores cross-chain interactions and does not include the auxiliary time prediction task.
        
   (ii) \textbf{w/ Group Mutual Excitation:} Built on the Base model, this variant adds the proposed \emph{Neural Group-wise Mutual Excitation} module to explicitly capture interactions across frequency-based groups, providing an efficient approximation to full-rank mutual excitation.
    
    (iii) \textbf{Full Model (w/ Time Pred -- NCQ):} The complete framework, which further incorporates the auxiliary time prediction task using our proposed \textbf{Non-Crossing Quantile (NCQ)} objective.
    
   (iv) \textbf{Variant (w/ Time Pred -- MSE):} A contrastive variant that replaces NCQ with standard mean squared error (MSE) for time prediction, used to assess the necessity of the quantile-based formulation.

\paragraph{Effect of group-wise mutual excitation.}
As illustrated in Figure~\ref{fig:ablation_study}, the group-wise mutual excitation mechanism is the primary driver of performance gains across all benchmarks. By comparing the \textbf{Base (attnhp)} model with the \textbf{gattnhp} variant, we observe a substantial leap in predictive accuracy. For instance, on \textit{ICEWS18}, the MRR rises from 0.2757 to 0.3956 (+11.99 percentage points), and on \textit{ICEWS14}, it increases from 0.4431 to 0.5138. 

These results empirically validate that events in temporal knowledge graphs are not isolated; rather, they are governed by complex cross-chain dependencies. While standard self-attention (Base) only captures historical self-excitation within a single dyad, our proposed grouping strategy allows event chains to borrow statistical strength from their latent semantic neighbors. This macro-level modeling effectively captures the hidden interaction topologies and alleviates the sparsity issues inherent in individual sequences, leading to more robust and comprehensive temporal representations.

\paragraph{NCQ vs. MSE for multi-task time prediction.}
We further investigate the impact of the auxiliary time prediction objective on the primary entity forecasting task. As shown in the comparison between \textbf{gattnhp+time+MLPCNQ} and \textbf{gattnhp+time+MLPMSE}, the choice of loss function is critical for stable multi-task learning. 

The proposed \textbf{NCQ} objective maintains, and in some cases further enhances, the entity prediction performance (e.g., achieving the highest MRR of 0.3863 on \textit{ICEWS18} and 0.5122 on \textit{ICEWS05-15}). This indicates that our quantile-based probabilistic modeling provides a compatible and enriching auxiliary signal that aligns well with the intensity-based event prediction. 

In sharp contrast, the \textbf{MSE} variant triggers a catastrophic performance collapse across all datasets. For example, on \textit{ICEWS18}, the MRR plummets from 0.3863 to a mere 0.2415, and on \textit{ICEWS14}, it drops from 0.5118 to 0.1124. This drastic degradation confirms our theoretical intuition: because inter-event times in TKGs follow heavy-tailed distributions with extreme outliers, the squared-error penalty of MSE generates excessively large, "toxic" gradients during backpropagation. These gradients dominate the optimization process and distort the shared latent representation space, effectively destroying the model's ability to reason about entity relationships. Consequently, our NCQ formulation is essential for achieving a successful joint optimization of "what" and "when" in TKG extrapolation.

{\subsubsection{Sensitivity to the number of latent groups ($G$).}
To investigate the scalability and effectiveness of our semantic soft grouping strategy, we evaluate the model's performance under varying numbers of latent groups $G \in \{1, 2, 4, 8, 16\}$. As shown in Figure~\ref{fig:group_sensitivity_all}, the choice of $G$ plays a critical role in balancing the primary entity prediction task (MRR) and the auxiliary time prediction task (MAE).

Setting $G=1$ corresponds to a configuration where all events are assigned to a single, unified group. In this scenario, the mutual excitation matrix effectively learns a global, macro-level interaction pattern across the entire knowledge graph. While this global prior provides a robust baseline and yields respectable performance (e.g., an MRR of 0.4565 on ICEWS14), it fundamentally lacks the capacity to capture fine-grained semantic topologies. As $G$ increases, the model explicitly disentangles diverse event patterns from the global average, enabling more precise and specialized mutual excitations. 

Interestingly, we observe a crucial trade-off between the primary entity prediction task (MRR/Hits@K) and the auxiliary time prediction task (MAE). For instance, while setting $G=2$ yields slightly higher MRR scores on some datasets, it suffers from a marked degradation in time prediction (e.g., MAE spikes significantly). Conversely, $G=4$ strikes the best balance across all three datasets, maintaining highly competitive MRR scores while substantially reducing the time prediction error. We therefore fix $G=4$ in all experiments.

However, excessively large values (e.g., $G=16$) consistently lead to a performance drop across both tasks. We attribute this to \emph{group-level sparsity}: over-partitioning the semantic space dilutes the statistical strength shared among event chains, hindering the Hawkes process from learning robust excitation priors.

\begin{figure*}[htbp] 
  \centering
  \includegraphics[width=0.6\linewidth]{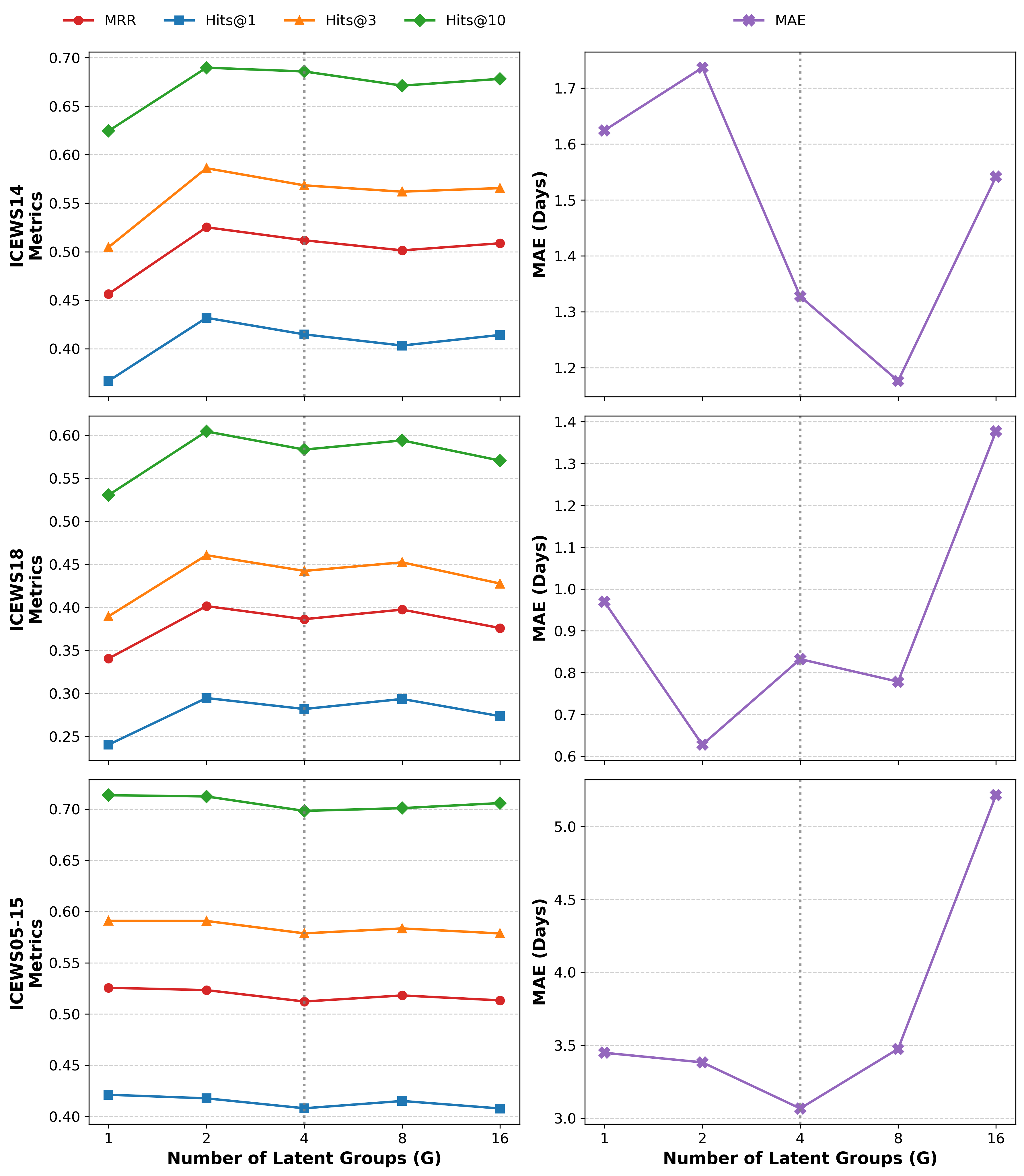}
  
   \caption{Sensitivity analysis of the latent group number ($G$) across three datasets. For each dataset, the left panel shows entity prediction performance (MRR and Hits@K, higher is better), and the right panel shows the time prediction error (MAE in days, lower is better). The vertical dotted line indicates our selected $G=4$, which strikes an optimal balance between the two forecasting tasks across all datasets.}
  \label{fig:group_sensitivity_all} 
\end{figure*}

\subsection{Why Group Interaction Matters: A Frequency-Aware Analysis}
\label{sec:why_group}

To isolate \emph{where} the performance gains of our proposed group interaction module originate, we conduct a fine-grained analysis by partitioning the test set into three categories based on the training frequency of each event chain $u=(s,r)$. Specifically, chains with training frequencies below the 90\textsuperscript{th} percentile are classified as \textit{Low-Freq (Tail)}, those above the 99\textsuperscript{th} percentile as \textit{High-Freq}, and the remainder as \textit{Mid-Freq}. We then compare our full model, \textbf{GAttNHP}, against its ablated variant, \textbf{AttNHP}, which retains the self-attention encoder but removes the cross-sequence mutual excitation entirely.

Table~\ref{tab:freq_breakdown} reports the MRR and Hits@$K$ for each frequency group.
The results reveal a striking pattern: the benefits of the group interaction module are \emph{not} uniformly distributed. While high-frequency chains already enjoy abundant training data and achieve reasonable performance without cross-sequence interaction, the most substantial gains consistently emerge in the \textbf{Mid-Freq} and \textbf{Low-Freq (Tail)} groups. For instance, on the tail group, GAttNHP achieves an MRR of 0.5415 and a Hits@10 of 0.6944, representing massive absolute improvements of 10.68 and 10.93 percentage points over AttNHP, respectively.

\begin{table}[htbp]
\centering
\caption{Performance breakdown by event chain frequency. \textbf{Bold} indicates the best result in each row. $\Delta$ denotes the absolute percentage point improvement of GAttNHP over AttNHP.}
\label{tab:freq_breakdown}
\begin{tabular}{lcccc}
\toprule
\textbf{Frequency Group} & \textbf{Metric} & \textbf{AttNHP (w/o Group)} & \textbf{GAttNHP (w/ Group)} & \textbf{$\Delta$} \\
\midrule
\multirow{4}{*}{High-Freq} 
    & MRR          & 0.4339 & \textbf{0.4693} & +3.5\% \\
    & Hits@1       & 0.3338 & \textbf{0.3602} & +2.6\% \\
    & Hits@3       & 0.4875 & \textbf{0.5285} & +4.1\% \\
    & Hits@10      & 0.6113 & \textbf{0.6683} & +5.7\% \\
\midrule
\multirow{4}{*}{Mid-Freq} 
    & MRR          & 0.4545 & \textbf{0.5402} & +8.6\% \\
    & Hits@1       & 0.3697 & \textbf{0.4486} & +7.9\% \\
    & Hits@3       & 0.5036 & \textbf{0.5999} & +9.6\% \\
    & Hits@10      & 0.6063 & \textbf{0.6988} & +9.3\% \\
\midrule
\multirow{4}{*}{Low-Freq (Tail)} 
    & MRR          & 0.4347 & \textbf{0.5415} & +10.7\% \\
    & Hits@1       & 0.3611 & \textbf{0.4601} & +9.9\% \\
    & Hits@3       & 0.4670 & \textbf{0.5833} & +11.6\% \\
    & Hits@10      & 0.5851 & \textbf{0.6944} & +10.9\% \\
\bottomrule
\end{tabular}
\end{table}

This phenomenon perfectly aligns with the intended mechanism of our semantic soft-grouping strategy. Low-frequency chains suffer from severe data sparsity, making it difficult to learn reliable temporal dynamics solely from their limited individual histories. Through the group-level mutual excitation module, these tail chains dynamically aggregate statistical strength from other chains within the same latent semantic group---specifically, chains that share similar subject-relation semantics and excitation patterns. In effect, the module acts as an \emph{implicit data augmentation} mechanism. Rare events are no longer modeled in isolation; instead, they benefit from the collective dynamics of their semantic peers.

In contrast, high-frequency chains gain only marginal improvements from the group module, as their own rich histories already provide sufficient training signals for the self-attention encoder. This explains why GAttNHP's overall MRR improvement is primarily driven by its superior predictions on mid- and tail-frequency events---precisely the long-tail entities that existing TKG models struggle with most.

\paragraph{Takeaway.} The group interaction module does not merely add capacity; through cross-chain group interaction, data-scarce chains borrow statistical strength from data-rich ones, which is precisely why its gains concentrate on the mid- and low-frequency chains.

\subsubsection{Case Study: The Impact of Group Interaction}
\label{sec:case_study}

To qualitatively understand how the group-level mutual excitation module corrects the bias of relying solely on individual event histories, we conduct a case study on two representative queries from the ICEWS14 dataset. 
We compare the top-3 ranking predictions of our full model, \textbf{GAttNHP}, against its ablated counterpart, \textbf{AttNHP} (which strictly relies on self-attention over isolated event chains). Table~\ref{tab:case_study} reports the predictions, with the ground-truth answers \underline{underlined}.

\begin{table}[htbp]
\centering
\caption{Comparison of top-3 predictions between AttNHP (without group module) and GAttNHP (with group module) on ICEWS14. The ground-truth entity is \underline{underlined}.}
\label{tab:case_study}
\resizebox{\linewidth}{!}{%
\begin{tabular}{clcll}
\toprule
& \textbf{Query $(s, r, ?, t)$} & \textbf{Rank} & \textbf{AttNHP (w/o Group)} & \textbf{GAttNHP (w/ Group)} \\
\midrule
\multirow{3}{*}{\textbf{Q1}} 
    & \multirow{3}{5.2cm}{\footnotesize (North Atlantic Treaty Organization, Consult, \underline{John Kerry}, 2014-11-01)} 
    & 1st & Yunus Qanuni & \underline{John Kerry} \\
    & & 2nd & \underline{John Kerry} & Barack Obama  \\
    & & 3rd & Afghanistan & Joseph\_Robinette Biden \\
\midrule
\multirow{3}{*}{\textbf{Q2}} 
    & \multirow{3}{5.2cm}{\footnotesize (Armed Gang (Syria), Use unconventional violence, \underline{Military (Lebanon)}, 2014-11-01)} 
    & 1st & Hezbollah & \underline{Military (Lebanon)} \\
    & & 2nd & \underline{Military (Lebanon)} & Terrorist (Syria) \\
    & & 3rd & Citizen (Australia) & Citizen (United Kingdom) \\
\bottomrule
\end{tabular}%
}
\end{table}

\paragraph{Discussion and Insights.} 
The results in Table~\ref{tab:case_study} perfectly illustrate the ``information borrowing'' capability of GAttNHP. 
In \textbf{Q1}, the query asks who NATO would consult in November 2014. AttNHP predicts Yunus Qanuni (an Afghan politician), likely overfitting to NATO's localized historical presence in Afghanistan. However, by incorporating group-level semantics, GAttNHP successfully captures the macro-level geopolitical alignment and correctly ranks \underline{John Kerry} (then US Secretary of State) at the top, even placing other relevant US figures (Barack Obama, Joe Biden) highly. 

Similarly, in \textbf{Q2}, an armed gang in Syria uses unconventional violence. AttNHP incorrectly predicts Hezbollah as the primary target, driven by raw historical frequency in the region. In contrast, GAttNHP successfully identifies the \underline{Lebanese Military}. This indicates that the group interaction module helps the model recognize the broader structural pattern of Syrian armed gangs spilling over to clash with neighboring state militaries, rather than memorizing a single, highly active entity like Hezbollah. Both cases demonstrate that macro-level group priors effectively prevent the model from being trapped in narrow historical biases.}

\end{document}